**Exploring the Limitations of kNN Noisy Feature Detection and Recovery for Self-Driving Labs**

Qiuyu Shi[1], Kangming Li[2], Yao Fehlis[3], Runze Zhang[1], Daniel Persaud[1], Robert Black[4], Jason Hattrick-Simpers[1,2,5,6,7*]

1. Department of Materials Science and Engineering, University of Toronto, Toronto, ON, Canada
2. Acceleration Consortium, University of Toronto, Toronto, ON, Canada
3. Artificial, Inc., Austin, Texas, United States
4. Clean Energy Innovation Research Center, National Research Council Canada, Mississauga, ON, Canada
5. Natural Resources Canada, Mississauga, ON, Canada
6. Vector Institute for Artificial Intelligence, Toronto, ON, Canada
7. Schwartz Reisman Institute for Technology and Society, Toronto, ON, Canada

[*]Corresponding author: jason.hattrick.simpers@utoronto.ca

## Abstract

Self-driving laboratories (SDLs) have shown promise to accelerate materials discovery by integrating machine learning with automated experimental platforms. However, errors in the capture of input parameters may corrupt the features used to model system performance, compromising current and future campaigns. This study develops an automated workflow to systematically detect noisy features, determine sample-feature pairings that can be corrected, and finally recover the correct feature values. A systematic study is then performed to examine how dataset size, noise intensity, noise type, and feature value distribution affect both the detectability and recoverability of noisy features on both Density Functional Theory (DFT) and SDL datasets. In general, high-intensity noise and large training datasets are conducive to the detection and correction of noisy features. Low-intensity noise reduces detection and recovery but can be compensated for by larger clean training data sets. Detection and correction results vary between features, with continuous and dispersed feature distributions showing greater recoverability compared to features with discrete or narrow distributions. This systematic study not only demonstrates a model agnostic framework for rational data recovery in the presence of noise, limited data, and differing feature distributions but also provides a tangible benchmark of kNN imputation in materials datasets. Ultimately, it aims to enhance data quality and experimental precision in automated materials discovery.

## 1. Introduction

Self-Driving Labs (SDLs) are revolutionizing scientific research and industrial processes[1–8]. By integrating robotics, artificial intelligence (AI), and advanced data analytics, these platforms are positioned to significantly boost productivity and reproducibility, promising rapid hypothesis testing and accelerated discovery cycles[8–12]. The automation of repetitive and complicated experimental tasks can help reduce the potential for human errors in materials investigation[1,10,12,13]. Moreover, AI-driven decision-making algorithms continuously learn from incoming data and recommend optimal experiments in real time, further enhancing efficiency, reducing human bias, and providing novel insights[10,12,14,15]. However, this premise relies on the consistency of the feature data collected by the platform, which is used to describe the experiment and is eventually input for the AI[6,12].

Noise or inconsistency in the data that an SDL is monitoring or producing propagates through the model and is a significant source of errors between the model's predictions and experimental outcomes[14,16–19]. Potential sources of noise could include equipment malfunctions, miscalibrations, and drift, as well as noisy data collected from characterizations[20]. For example, during material synthesis via physical vapor deposition, vacuum gauge calibration drift may provide a distorted relationship between deposition back pressure and the synthesis of a desired phase[21–23]. A further complication is that during data pre-processing and cleaning, the feature data is often normalized which can make noise detection challenging. Therefore, the development of a framework and clear guidelines for when it is possible to identify and correct noisy features is critical for the rational and successful broader deployment of SDLs[14].

To address noisy features, both imputation and correction techniques can be applied. Simple statistical imputation, such as filling in missing values with the mean, median, or mode, offers a quick and easy solution but often fails to capture relationships between variables[24,25]. Regression-based imputation and Multivariate Imputation by Chained Equations (MICE) model missing values as functions of other variables, which can make them more accurate in complex datasets[26,27]. For correcting noisy features, methods like outlier detection (e.g., Z-score, Isolation Forest) and feature transformation (e.g., log scaling, binning) help stabilize variance and reduce distortion[28–30]. Finally, denoising autoencoders, robust regressors, and dimensionality reduction techniques (like PCA) not only impute values but also correct underlying data noise[31–33]. Within the materials informatics field, k-Nearest Neighbors (kNN) is a widely applied imputation and correction method[34–36]. It fills in missing or noisy values using the values of the nearest training points, based on a distance metric (e.g., Manhattan, Euclidean). However, to the best of our knowledge, no systematic study exists that comprehensively evaluates the performance of the kNN method under various intensities of noise, training dataset size and feature distributions.

In this work, we addressed this gap by focusing on both the detection and recovery performance of kNN method on noisy features with both computational and real SDL materials science datasets, as well as systematically exploring their limitations across varying noise types, noise intensities, and training dataset sizes. We present a comprehensive workflow for noisy feature detection and recovery and investigate its performance under diverse noise scenarios. Specifically, we analyze how additive Gaussian, Poisson and Drift noise affects feature detection and recovery. By simulating instrumental errors through the introduction of systematic noise to certain features, we assess the model's capability to detect and correct these deviations. Moreover, we evaluate the robustness of our detection methods by adjusting the noise magnitude to explore a range of signal-to-noise ratios, including conditions where noise levels rival or exceed the true signal. Additionally, we examine the impact of training dataset size on model performance, investigating whether smaller datasets increase susceptibility to overfitting and impair noise detection. Conclusively, our study provides valuable insights and practical guidance for mitigating overly optimistic imputation and correction. It provides a framework for systematically interrogating the feasibility and accuracy of feature detection and correction, which can be used to improve data quality in SDLs and thereby enhance the reliability and performance of automated experimental systems.

## 2. Methodology

In this study, we employ a system validation framework in two distinct datasets: a computational DFT dataset representing simulated materials properties, and a 3D printing SDL dataset capturing experimental manufacturing variables. Moreover, to evaluate the applicability of the model under different noisy data scenarios, three noise introduction methods were used to simulate experimental uncertainties in the real world. Afterwards, this section ends with a four-step explanation of the full workflow, covering data preparation, baseline model evaluation, noisy feature detection, recoverable sample determination and corrections.

### 2.1 Dataset Selection

Two datasets were employed for training, validation, and testing the workflow. A computational dataset was selected (JARVIS-DFT formation energy) contains 71,571 data points with 273 compositional and structural features and was extracted via Matminer using Jarvis-tools[38,39]. To simulate realistic dataset characteristics collected and analyzed from SDLs, an initial feature elimination process was applied to the dataset. First, highly correlated features were removed using a Pearson correlation threshold of 0.7, leaving 88 features[40,41]. Next, to further refine the feature set based on their relevance to the formation energy target, two machine learning models, Random Forest and XGBoost, were trained[42]. Each model was trained on 80% of the dataset and tested on the remaining data. Features were ranked using the Gini impurity-based importance method, and those contributing to a cumulative feature importance of 0.9 were selected[43]. The union of important features from both models resulted in 46 total features. The feature value distribution plots for each feature can be found in Figure S1.

The other publicly available dataset is from 3D printing SDL experiments to further evaluate the applicability of our workflow in real-world scenarios[44]. This dataset comprises 13,250 experimental runs collected over two years using a platform equipped with five fused filament fabrication 3D printers, a precision scale, and an Instron mechanical tester[44]. The dataset includes 12 features representing various process parameters and physical measurements recorded throughout the SDL experiments. Additional information about the feature descriptions, their correlations, and the feature distributions can be found in Figures S2 and S3, respectively.

### 2.2 Noise Introduction Methods

Three types of noise were tested with the workflow: Gaussian, Poisson and Drift noise. Gaussian noise is a common type of noise in measurement systems, representing the thermal noise in electronics and background fluctuations, and is characterized by a normal probability distribution[45]. It was introduced by adding random values sampled from a zero-mean normal distribution onto the input data [46]. Poisson noise typically exists in physical instrumentation that records signals from discrete events, such as electron transfer processes in electrochemical impedance spectroscopy (EIS)[47]. It was modeled by first determining a rate parameter based on the absolute magnitude of the data, and then generating Poisson-distributed random counts, which were subsequently added to the input data[48]. Drift noise reflects the gradual accumulation of measurement error over time, often caused by sensor aging or environmental fluctuations. The Drift noise was introduced using a random walk model, where each step was a random variable sampled from a normal distribution and accumulated over time using a cumulative sum along the time dimension. [49,50].

### 2.3 Workflow Overview

The overall workflow includes four steps: baseline model evaluations, noisy feature detection, recoverable sample determination, and recoverable noisy sample corrections, as illustrated in Figure 1. Before entering the four steps, the data were divided into training, validation, and test sets using an 8:1:1 split. Each subset was then min-max scaled separately to the [0, 1] interval, ensuring that scaling parameters derived from one subset did not leak information into the others[51].

#### 2.3.1 Baseline Model Evaluations

Before applying the detection or correction methods to noisy feature data, the kNN imputation model's accuracy was validated through a systematic hyperparameter tuning process using *GridSearchCV* with 5-fold cross-validation[52]. This optimization phase identified a kd-tree search algorithm with a neighborhood size of five, which was determined to provide the optimal balance between local sensitivity and noise robustness. To maximize tree traversal efficiency in high-dimensional space, a leaf size of 30 was selected. Furthermore, the Manhattan (L1) metric (p=1) was identified as the most effective distance measure, with predictions weighted inversely by distance to emphasize the influence of proximal neighbors. Once established, these optimized hyperparameters were held fixed across all subsequent experiments.

Each feature was sequentially treated as the target (the $N^{th}$ feature) and imputed using the remaining N-1 features from the training set. The difference between the recovered and original values was then calculated and labelled as Δbase, which serves as a reference for the noisy feature detection step. Separate kNN models were used for each target feature during this recovery process and tested with five random seeds. To evaluate the reliability of these recovery results, the coefficient of determination ($R^2$) of each model was calculated and summarized in Figure S4, providing a measure of how well the imputed values matched the original JARVIS-DFT data[36,53].

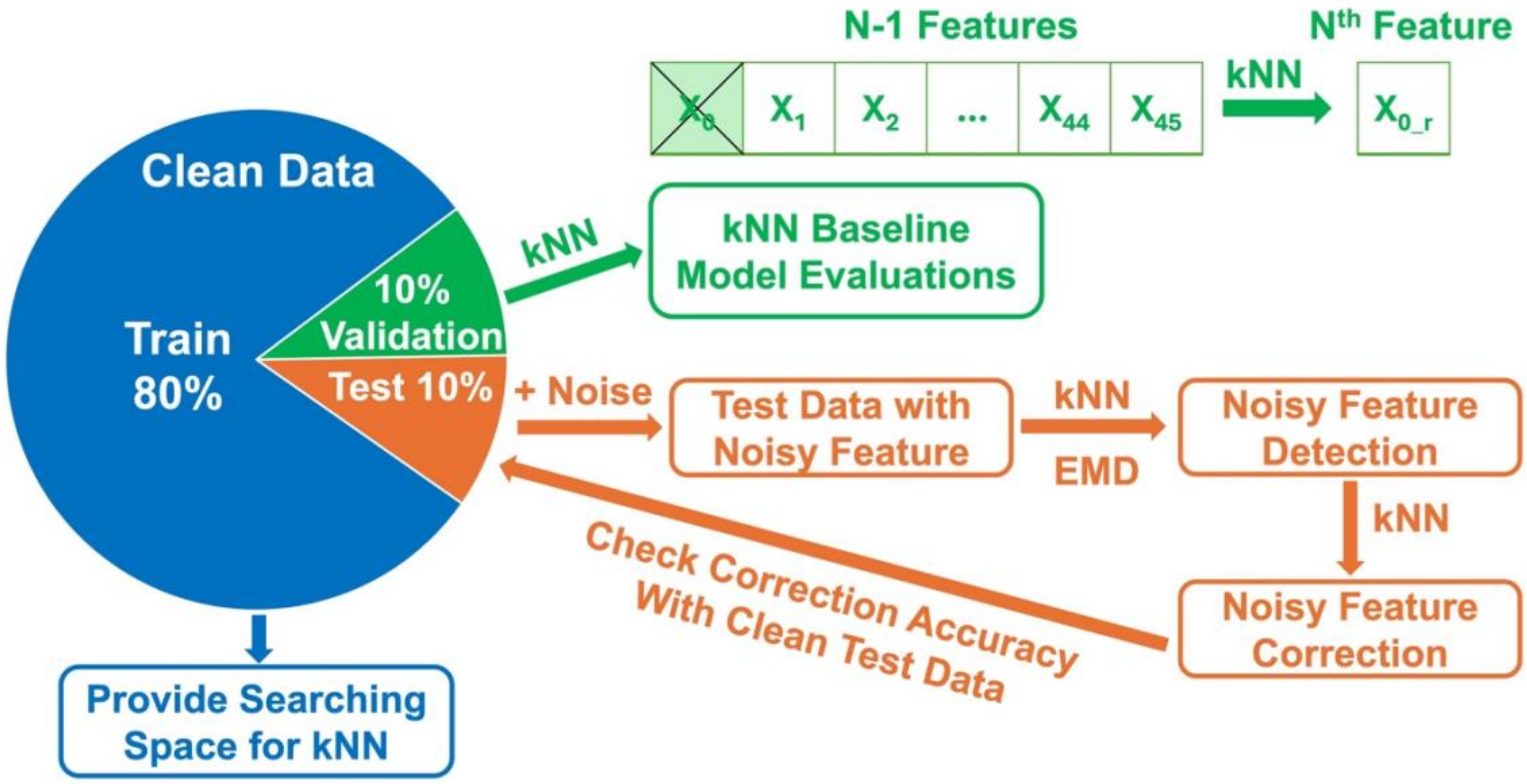

**Figure 1. Overview of the noisy feature detection and correction workflow.**

### 2.3.2 Noisy Feature Detection

The noisy test set is first obtained by introducing one type of noise into one feature at a time. For the detection step, a collaborative approach employing kNN imputation and Earth Mover's Distance (EMD) was implemented[54–57]. Under the assumption that there is a feature with unknown noise in the test set, the baseline evaluation process is applied sequentially to each feature with kNN, treating each one in turn as if it were the noisy feature and the recovery is performed individually. For each feature, the difference between the recovered value and the original value in the test set is calculated, resulting in a data frame of differences for all features, denoted as $\Delta$noise.

To evaluate these recovery results relative to the baseline, we compare $\Delta$noise with $\Delta$base by plotting both distributions together on a violin plot separated by a black line, as shown in Figure 2 (a). To quantitatively assess the similarity between distribution pair, we employed the EMD method. This method calculates the minimum cost required to transform one distribution into another, providing an accurate measure of dissimilarity, particularly for low-dimensional data. [54,58]. For each noisy feature scenario, the feature with the largest EMD value between $\Delta$base and $\Delta$noise is identified as the noisy feature. For instance, Figure 2 (a) presents distribution pairs from four example features, showing *MagpieData Minimum GSbandgap* feature has the largest EMD between all features, which leads to its identification as the noisy feature.

If this detected noisy feature matches the one to which noise was introduced in the previous noise introduction step, it is counted as a successful detection. Repeating this entire process for all features allows for the calculation of the overall successful detection rate, referred to here as detectability. After, the detectability was calculated under different noise intensities, training dataset size and noise types.

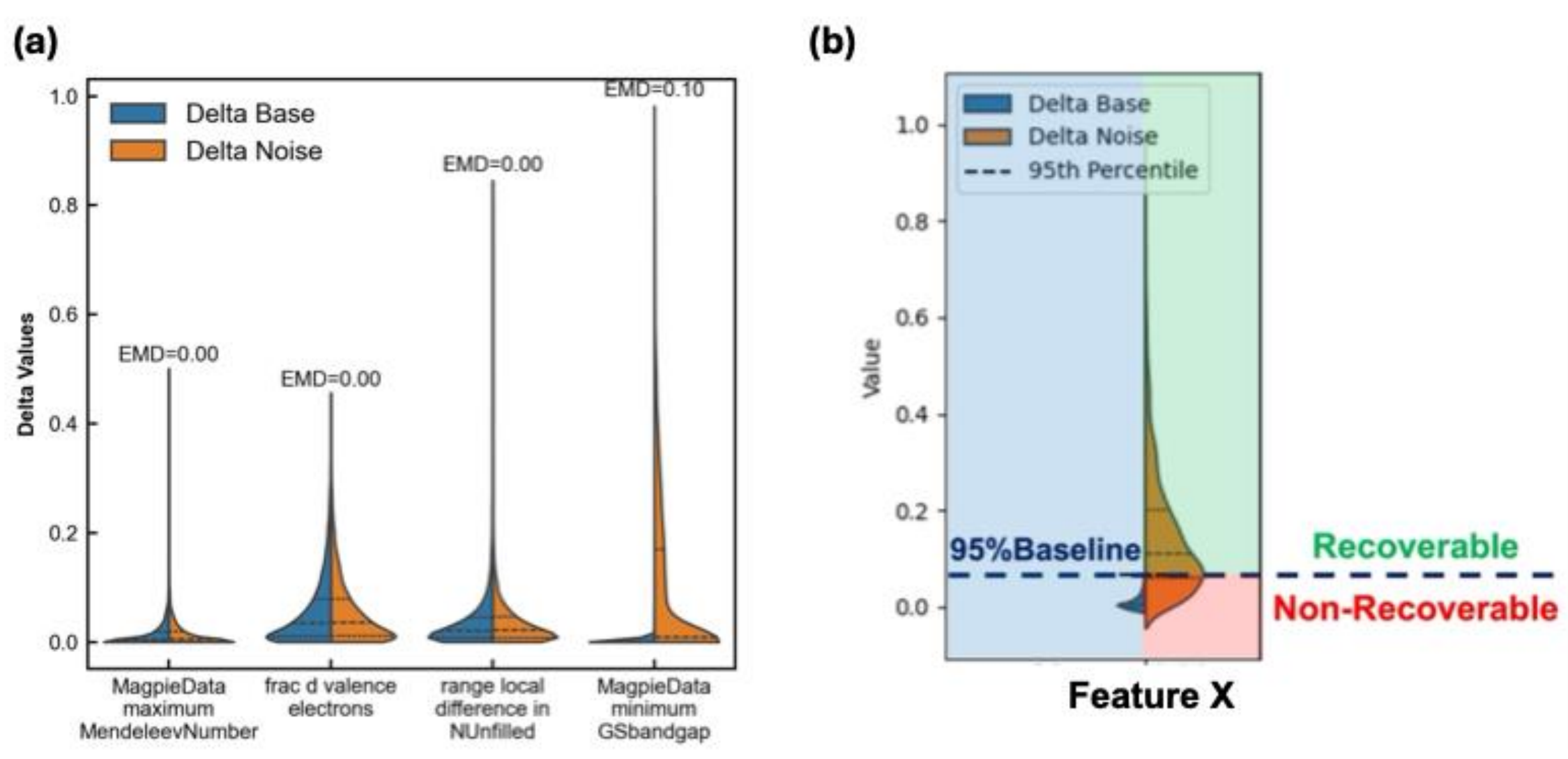

**Figure 2. Noise detection and recoverability criteria.** (a) Example detection of noise in the "*MagpieData Minimum GSbandgap*" feature. (b) Definition of the recoverability through the evaluation of baseline and recovery error distributions.

### 2.3.3 Recoverable Sample Determination

After identifying the noisy features, we determined which specific samples required correction based on the recoverability criteria illustrated in Figure 2 (b). Following the same format as the noise detection step, the baseline error distribution (Δbase) is plotted on the left, but with a dashed line marking the $95^{th}$ percentile. The recovery error distribution (Δnoise) is shown on the right. A sample is classified as recoverable, and thus targeted for correction, if its Δnoise exceeds this $95^{th}$ percentile threshold of the baseline error. The recoverability metric is then defined as the ratio of recoverable samples to the total number of samples in the test set. Using this workflow, we investigated the recoverability of various samples across a range of noise intensities.

### 2.3.4 Recoverable Noisy Sample Corrections

After determining the recoverable samples, the correction of noisy features could be done using the N-1 features to correct the $N^{th}$ feature. The correction accuracy was evaluated by comparing the imputed value for all recoverable samples to the ground truth value.

## 3. Results

### 3.1 kNN Baseline Model Accuracy Evaluations

The baseline model accuracy was first evaluated with the JARVIS-DFT dataset. Figure 3 (a) summarizes the $R^2$ values for 10 example features across varying training data sizes, ranging from 0.1k to 57k samples. The x-axis represents the target features, while the y-axis displays the corresponding $R^2$ values. Solid points indicate the mean $R^2$ values obtained from five different random seed experiments, and the surrounding bands represent the standard deviation. In general, most features achieve an $R^2$ above 0.8 when using the full-size training set, reflecting a relatively high prediction accuracy and laying a foundation for the subsequent noise detection and recovery steps. As the training data size decreases, a corresponding drop in $R^2$ values is observed. This trend is consistent with the kNN model mechanism: larger training sets provide a more concentrated pool of neighbors, thereby enhancing prediction accuracy[34].

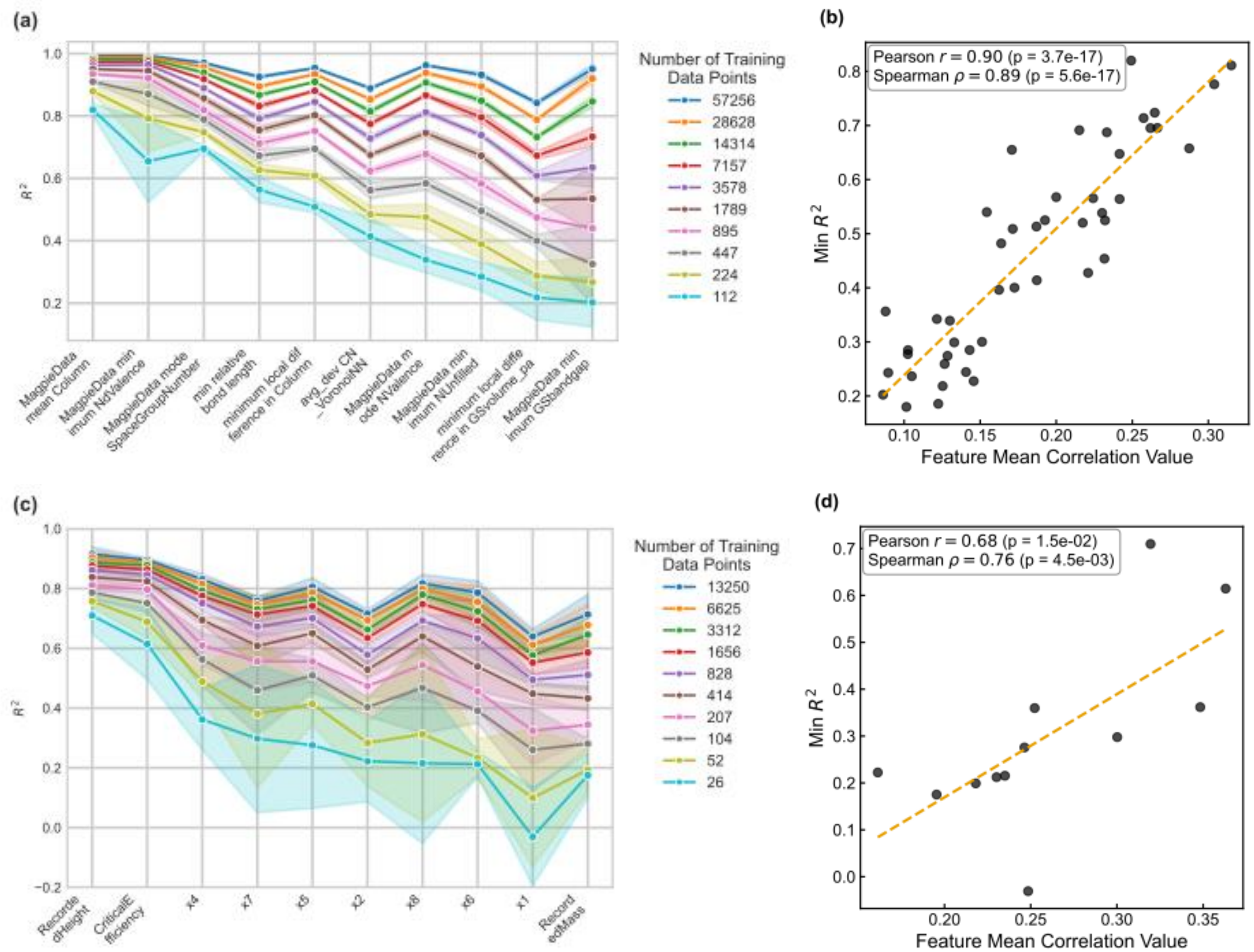

**Figure 3. Baseline accuracy and feature correlation analysis.** $R^2$ performance for ten kNN models across training volumes in (a) JARVIS-DFT and (c) SDL dataset. (b) and (d) illustrate the relationship between model $R^2$ and feature mean correlation at the minimum training sizes investigated for each respective dataset.

A closer examination of Figure 3 (a) reveals that the performance of the model depends strongly on which feature is recovered. Notably, some features, such as the *MagpieData mean Column*, can maintain relatively high $R^2$ scores even with significant reductions in training data, suggesting that these features are inherently more robust and less sensitive to data scarcity. Conversely, some features, such as the *minimum local difference in GSbandgap*, exhibit a wider range of $R^2$ values across different training sizes, indicating a higher sensitivity to the amount of available data. Since the kNN correction method is based on the correlations between features, here we investigated the relationship between feature correlations and the $R^2$ value with the smallest training dataset size. A strong linear correlation between the mean of feature correlations and its corresponding $R^2$ was found and plotted in Figure 2 (b). This plot indicates that the greater the average correlation between the feature being corrected and the other features, the higher its correction accuracy, which aligns with the kNN correction method's reliance on neighbor-based similarity[35].

The same test was performed on the 3D printing SDL dataset. From the results summarized in Figure 3 (c). We observed that the baseline model has an $R^2$ value above 70% with training data sizes larger than 10k, but performance degrades significantly as the number of training samples decreases. This trend is consistent with what we observed in the JARVIS-DFT dataset. However, the reduced number of features in this dataset leads to lower correction accuracy, as fewer and weaker correlated features are available for kNN to predict, reflected in the lower feature correlations seen in Figure 3 (d).

This baseline model study provides an initial applicability evaluation of the kNN model under varying levels of training data availability, offering researchers a practical reference for deploying this method in their own settings.

### 3.2 Noisy Feature Detection

After validating the accuracy of the kNN baseline model, we applied our proposed technique to detect noisy features using an example of a Gaussian noise scenario. Specifically, to simulate the meter precision errors commonly found in SDLs, Gaussian noise was added to a selected feature across all samples in the test set, generating a noisy test dataset. The results are summarized in Figure 4 for two datasets.

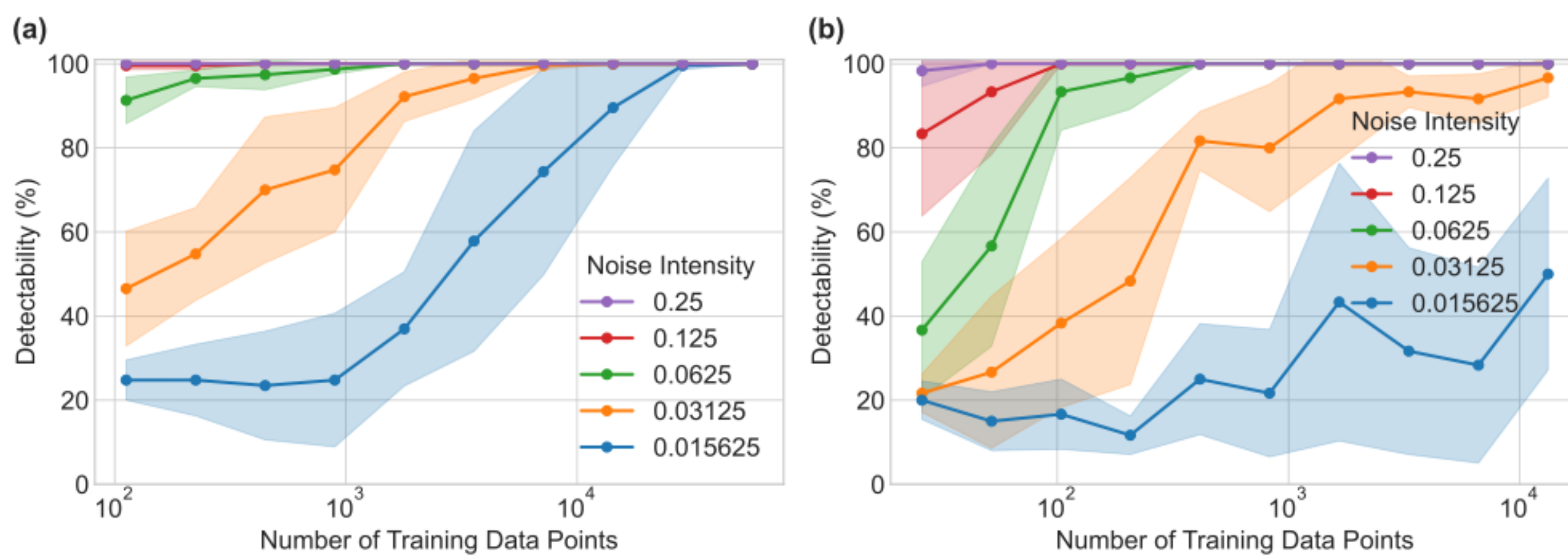

**Figure 4. Detectability summary as a function of noise intensities (σ) and training set sizes (N).** (a) JARVIS-DFT dataset results for $112 \leq N \leq 57,256$. (b) SDL dataset results for $26 \leq N \leq 13,250$. Noise intensity ranges from 0.015625 to 0.25 for both cases.

In the JARVIS-DFT dataset, we perform this study on different Gaussian noise intensities ($0.015625 \leq \sigma \leq 0.25$) and training dataset sizes (112 to 57256 points) with 5 random seeds to explore the detection limits under various scenarios. The detectability results are summarized in Figure 4 (a), with the mean values from 5 random seeds being the solid points and the standard deviation being the error bar range. In general, detectability increases with both the size of the training data and the intensity of the noise. With higher Gaussian noise intensity ($\sigma > 0.03125$) introduced in the test dataset, detectability could still be preserved above 80% even with limited amounts of training data. However, as the noise intensity continues dropping, the training dataset size becomes more critical. When the training data is less than 1,000 points, the model struggles to detect low-intensity Gaussian noise ($\sigma < 0.03125$). While such low noise levels often fall within instrument precision and may not require correction in real-world SDLs, we include this regime to benchmark the lower bounds of detectability and assess the robustness of our method across the full noise intensity spectrum. If detection of such low-intensity noise is truly required, improving sensitivity at this regime would likely need a larger training set to enhance the model's accuracy.

In the SDL dataset, the noisy feature detectability under varying intensities and training dataset sizes is also presented in Figure 4 (b). Similar to the DFT data, detectability remains robust at noise levels above 0.0625 with training sizes above 100 samples. However, for lower intensity Gaussian noise smaller than 0.0625, larger training sets are required to maintain acceptable detection accuracy.

The above results from the two datasets reveal that performance scales positively with both training data volume and noise intensity. With more than 100 training data points, the detectability remains robust and typically exceeding 80% for noise with intensities above 0.0625. Besides, the model's sensitivity to low-intensity noise ($\sigma < 0.03125$) is highly dependent on larger dataset sizes. Although these lower noise levels may approach instrument precision limits in real-world SDLs, evaluating this regime establishes the method's lower sensitivity bounds and demonstrates its overall robustness across the noise spectrum.

### 3.3 Recoverable Noisy Samples Determination

To quantitatively explore how recoverability varies under different noise types and intensities, as well as to provide constructive insights for similar studies, we applied the same analysis method while systematically varying the noise intensity parameters into two datasets, respectively.

For the JARVIS-DFT dataset, Figure 5 (a) summarizes three example features that exhibit distinct responses to changes in noise intensity, while Figure 5 (c) displays the value distributions of each feature before and after introducing Gaussian noise. Under high-intensity Gaussian noise of 0.25, the minimum feature recoverability exceeds 60%, meaning that 60% of the samples with a given noisy feature can be easily recovered. The remaining 40% of noisy samples are within the baseline correction error and cannot be recovered in downstream analyses. As noise intensity decreases, various features show different behaviors in response to noise intensity changes. The features characterized by broadly distributed values demonstrated higher sensitivity to variations in noise intensity, such as *MagpieData maximum MendeleevNumber*. In contrast, features with narrow distributions, such as *MagpieData minimum GSbandgap*, showed limited sensitivity to noise changes. Therefore, the feature with a more continuous value distribution is expected to experience stronger changes concerning noise intensity compared to a narrower distribution, underlying the importance of feature selection and examination when building prediction models.

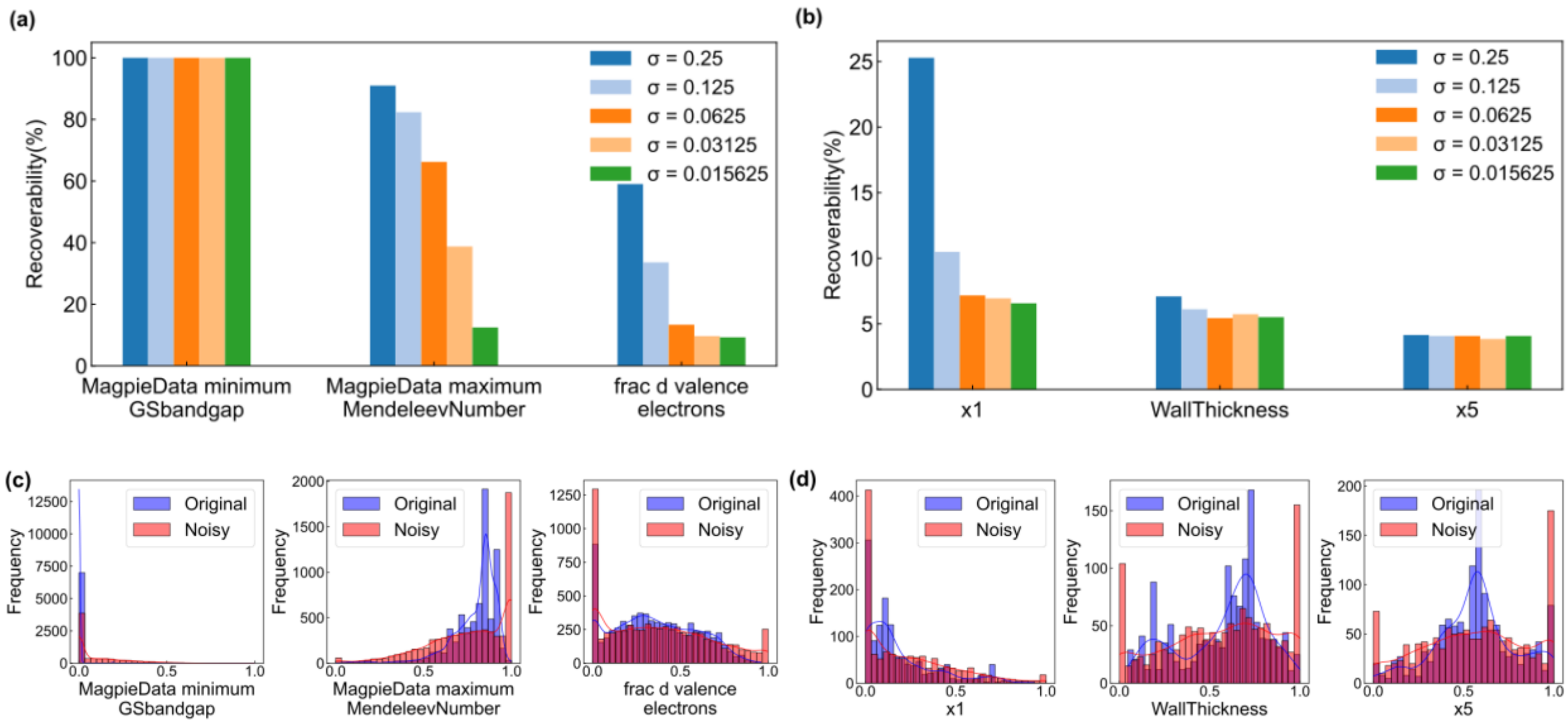

**Figure 5. Recoverability analysis and distribution shifts under varying noise intensities.** (a, b) Recoverability results for representative noisy features in the JARVIS-DFT and SDL datasets across Gaussian noise intensities ($0.0078125 \leq \sigma \leq 0.25$). (c) and (d) display the corresponding feature value distributions before and after the introduction of noise for the JARVIS-DFT and SDL datasets, respectively.

For the SDL dataset, examples of the recoverability results were shown in Figure 5 (b). Most features exhibit a recoverability rate near 5%, indicating that only a small fraction of samples can be successfully corrected. This outcome is influenced by our definition of recoverability, which is based on the divergence between the baseline error distribution (Δbase) and the post-noise distribution (Δnoise). When Δbase and Δnoise are nearly identical, few samples exceed the $95^{th}$ percentile threshold of Δbase, resulting in low recoverability.

A detailed examination of two representative features is presented in Figure 6. The feature "*MagpieData maximum MendeleevNumber*" from the JARVIS-DFT dataset exhibits a substantial divergence between Δbase and Δnoise. Consequently, a large proportion of samples in the Δnoise distribution fall above the $95^{th}$ percentile threshold, making them recoverable. In contrast, feature "x1" from the SDL dataset shows a much narrower gap between the two error distributions, resulting in a significantly lower quantity of recoverable samples.

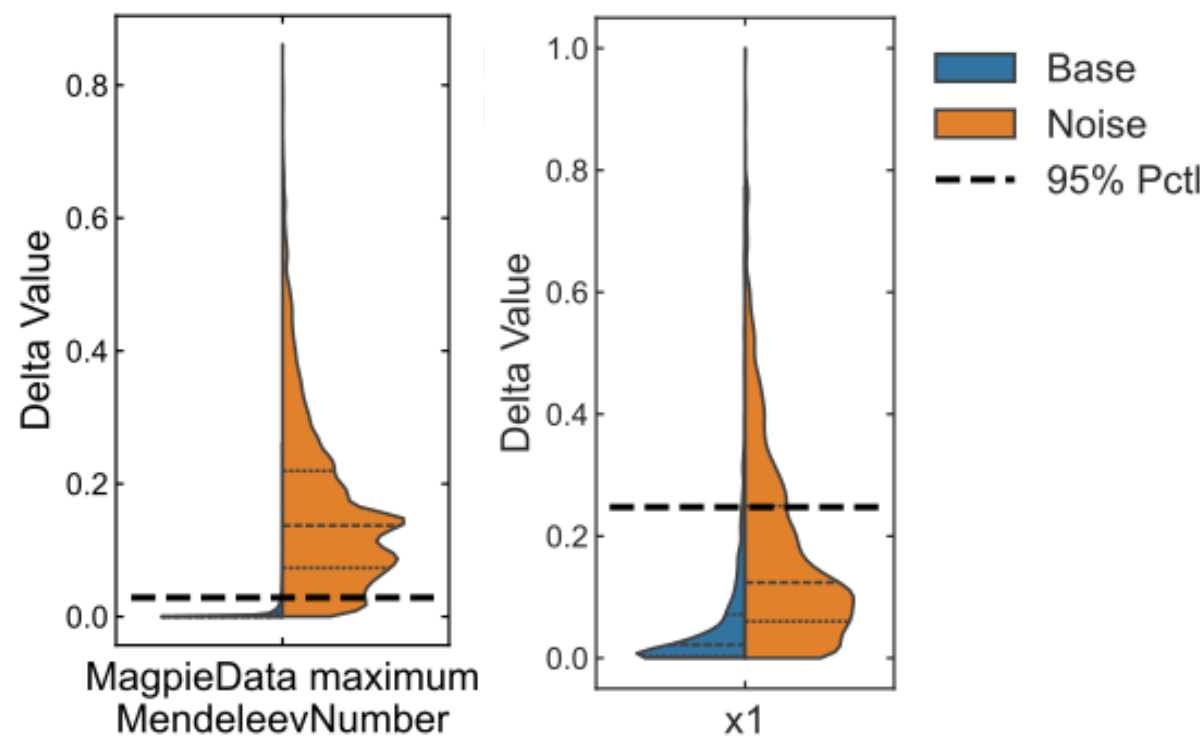

**Figure 6. Comparative analysis of sample recoverability for representative features.** Contrast between (a) *MagpieData maximum MendeleevNumber* (JARVIS-DFT), showing high recoverability due to a significant shift in relative to the 95$^{th}$ percentile threshold, and (b) feature *x1* (SDL), exhibiting minimal distribution divergence and lower recoverability.

The above recoverability analysis demonstrates that the underlying feature distribution before and after introducing noise plays a critical role in determining how many noisy samples are recoverable after detection. Features with broad, continuous distributions show significantly high recoverability rates exceeding 60% in the DFT dataset. In contrast, features with narrow distributions, particularly in the SDL dataset, exhibit minimal divergence from the baseline error, resulting in low recoverability as few samples exceed the 95$^{th}$ percentile threshold. These findings underscore the necessity of feature-specific examination when assessing model robustness to noise.

### 3.4 Noisy Feature Correction on Recoverable Samples

Once recoverable noisy features are identified, the subsequent step involves applying the kNN imputation method to recover these values. To evaluate the overall accuracy of the recovery method, each feature was treated as the noisy feature in turn, and the recovery accuracy was summarized in Figure 7.

For the JARVIS-DFT dataset, Figure 7 (a) provides an overview of the Mean Absolute Percentage Error (MAPE) of the recovered and original values for all features. Figure 7 (b) highlights an example feature with a high recovery accuracy. Overall, 85% of the recoverable noisy samples exhibit a high correction accuracy with an MAPE value under 20%. However, the correction accuracy varies considerably across features. Using the same 20% MAPE threshold, the proportion of accurately corrected samples varies by feature from 76.5% to 99.2%, highlighting the feature-dependent nature of recovery performance.

This variation in recovery performance appears to be closely linked to the underlying feature distribution. For example, Figure S6 (c) and (d) illustrate the distributions for two specific features: the *mean local difference in Electronegativity* and *MagpieData range NdValence*. The *Electronegativity* feature, which exhibits a more continuous distribution, is associated with more accurate recovery, whereas the *NdValence* feature, with its sparser distribution due to a limited set of possible values, shows larger discrepancies between the recovered and original values.

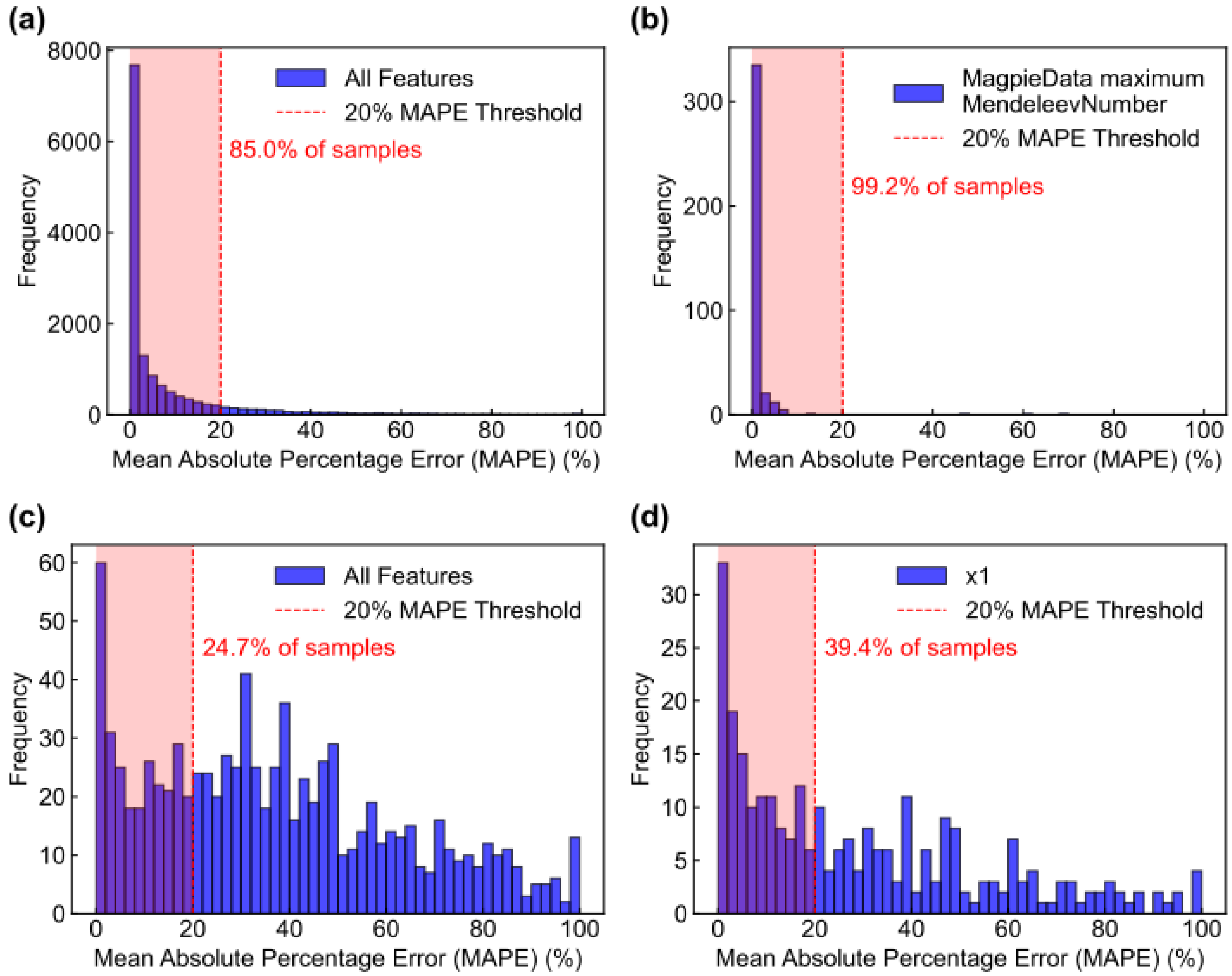

**Figure 7. Evaluation of correction accuracy for recoverable noisy features.** Global correction performance across all features for the (a) JARVIS-DFT and (c) SDL datasets. Detailed performance for the features exhibiting the highest correction accuracy in each dataset is presented in (b) and (d), respectively.

In the SDL dataset, the correction accuracy across features is summarized in Figure 7(c). Approximately 24.7% of samples achieve an MAPE below 20%. An example feature with high correction accuracy of 39.4% is shown in Figure 7(d). From the observation of another two example features shown in S7, the extent of correction varies by feature, with broader and more continuous feature distributions exhibiting better performance, while narrow or discretized distributions tend to result in lower accuracy. Compared to the earlier JARVIS-DFT dataset, the overall correction performance is reduced, likely due to the smaller feature set, weaker correlations, and smaller dataset size.

These findings demonstrate that recovery performance is strongly influenced by the underlying feature distributions. More broad and continuous distributions can yield better recoverability, while narrow or discrete distributions pose greater challenges. For the practical deployment of SDLs, this highlights the importance of carefully examining feature selection. Rather than discarding all correlated features to reduce redundancy, researchers should strategically retain them if the SDL system is prone to noisy data collection. In such scenarios, these correlations provide the essential context required for robust feature recovery and overall model stability.

### 3.5 Generalization of Detection and Correction to Alternative Noise Types

To further investigate the generalization of this workflow to different noise scenarios, here we did similar tests using the JARVIS-DFT dataset against another two types of noise: Poisson and Drift noise. Their detectability results, shown in Figure S8, have a similar trend compared with the Gaussian noise in Figure 4 (a). With a large noise intensity above 0.125 (12.5%), the detectabilities all preserve a high detection rate above 90% under different numbers of training datapoints ranging from 0.1k to 50k, showing the applicability of this workflow in noise detection function with large noise existing in the dataset. As the noise gradually decreases to below 0.0625 (6.25%), the number of training data points plays a more important role in noise detection. If we would like to obtain an accurate detection rate above 80% for noise with intensity below 0.03125 (3.125%), a larger amount of training data above 10k is needed. However, for the practical application scenario, 5% error window is commonly expected and tolerated[20]. Therefore, the results show us the applicability of this noise detection workflow on different noise types existing in SDL experimentations.

The recoverability results for Poisson and Drift noise are shown in Figure S9. Notably, the recoverability varied significantly between noise types. This variation is partly attributable to how recoverability is defined. Specifically, the proportion of samples in the noise correction distribution (Δnoise) exceeds the $95^{th}$ percentile of the baseline distribution (Δbase). When the difference between the baseline and correction values is small, the overlap between Δbase and Δnoise is large, resulting in fewer samples being identified as recoverable. Conversely, if the baseline model itself has larger correction errors, the $95^{th}$ percentile of Δbase is elevated, which also reduces the fraction of recoverable samples in Δnoise. Therefore, recoverability is influenced by both the nature of the noise and the statistical characteristics of the features involved.

Therefore, we extended the evaluation of the proposed noise detection and recovery workflow to more realistic SDL scenarios by introducing Poisson and Drift noise. The workflow demonstrated high detection accuracy across noise types, sustaining over 90% accuracy at higher noise levels and showing greater sensitivity to training dataset size at lower intensities. Recoverability results, however, were more variable and closely tied to both the type of noise and the distribution of feature values. These findings highlight the robustness of the detection strategy, while also emphasizing the importance of feature distribution and noise properties in determining recovery performance. Overall, the workflow shows strong potential for identifying noisy features under diverse conditions typical of experimental SDL environments.

# 4. Discussion

## 4.1 The Interplay of Training Dataset Size and Noise Intensity

Our evaluation of the kNN-based workflow reveals a fundamental trade-off between training dataset size and noise detection sensitivity. Consistent with the mechanics of kNN algorithms, which rely on the density of the neighbor pool, we observed that performance scales positively with dataset size. While high-intensity noise (e.g., $\sigma > 0.0625$) can be robustly detected over 80% even with limited data, detecting subtle noise near instrument precision limits (e.g., $\sigma < 0.03125$) requires a significantly larger training pool exceeding 10k samples. This suggests that for early-stage SDLs where data is scarce, the method is best suited for flagging catastrophic sensor failures rather than minor calibration drifts. However, as the experimental database grows, the system's sensitivity naturally improves, allowing for the correction of increasingly fine-grained errors.

## 4.2 The Role of Feature Distribution and Correlations

A critical finding of this study is that feature recoverability is not uniform. It is intrinsically linked to the statistical characteristics of the features themselves. We observed a strong linear correlation between the feature's baseline correction accuracy and its mean correlation with other features in the dataset. This confirms that the kNN approach relies heavily on the information provided by the neighbors.

Furthermore, the shape of the feature distribution plays a decisive role. Features with broad, continuous distributions exhibit high recoverability because noise introduces a clear divergence from the baseline distribution. In contrast, features with discrete or narrow distributions pose a challenge. The noise often fails to push the sample values beyond the $95^{th}$ percentile error threshold of the baseline, resulting in low recoverability rates. This highlights a practical imperative for SDL design: rather than aggressively pruning correlated features to reduce dimensionality, researchers should strategically retain them. These correlations provide the essential scaffolding required for robust error detection and recovery.

## 4.3 Generalizability Across Noise Types

The validation of our workflow against Gaussian, Poisson, and Drift noise demonstrates the robustness of the method across diverse experimental scenarios. While detection rates remained consistently high for high-intensity noise regardless of the noise type, the recoverability varied. This variation arises because different noise types alter the feature value distributions in unique ways, affecting the overlap between the clean and noisy error profiles. Therefore, while the system is highly reliable in alerting researchers to the presence of noise, its ability to correct noise depends on the specific statistical characteristics of the error source.

## 4.4 Implications for SDL Deployment

Overall, these results establish the limitations for the applicability of kNN-based noise correction. The method is highly effective for detecting and correcting significant noisy data in data-rich environments. For practical deployment, this implies that noise correction modules should be dynamic, activating different sensitivity thresholds based on the current volume of accumulated experimental data.

## 5. Conclusion

In summary, we have developed and validated a robust workflow for noise detection and recovery for SDLs. By combining clean datasets alongside a carefully designed feature elimination and selection process, we demonstrated that the kNN imputation method can effectively recover part of the noisy feature values in the presence of diverse intensity noise. Our analysis showed that detection and recovery accuracy depend critically on noise intensity, training dataset size and the inherent statistical distribution of the feature values. Noise intensity itself remains a key factor of performance across scenarios. A larger training dataset could compensate for the noise intensity and feature values. Features with broader distributions tend to be more recoverable, while narrowly ranged features exhibit limited recoverability. Moreover, the introduction of Δbase, Δnoise, and EMD metrics provides quantitative frameworks for detecting and quantifying noise, enabling precise identification of noisy features.

Overall, these findings not only advance real-time data-quality monitoring and troubleshooting in SDL, but also offer actionable guidance for researchers working with varying noise levels and dataset availabilities. This workflow development provides the community with a pipeline to test and evaluate the noise detection and correction methods under various noise intensities and dataset sizes, which can not only be kNN but also for other methods. Integrating our noise detection and recovery strategies into existing data management pipelines can substantially enhance the robustness, precision, and overall reliability of SDLs.

## Data Availability

The data used in the paper is accessible through the Zenodo repository at
https://zenodo.org/records/7659269

## Code Availability

The code for ML training, analysis, and figure generation in this work can be found at GitHub: https://github.com/SaraQiuyuShi/knn_noise_sdl

## Acknowledgements:

This research was undertaken thanks in part to funding provided to the University of Toronto's Acceleration Consortium from the Canada First Research Excellence Fund (Grant number: CFREF-2022-00042) and the National Research Council Canada (Materials for Clean Fuels Challenge-127).

**Declaration of Generative AI Tools**

The authors used ChatGPT (OpenAI) to assist with polishing the language, improving clarity, and refining grammar in parts of this manuscript. All content was critically reviewed and finalized by the authors to ensure technical accuracy and integrity.

**Conflict of interest:**

The authors declare no conflict of interest.